\def\year{2022}\relax
\documentclass[letterpaper]{article} 
\usepackage{aaai22}  
\usepackage{times}  
\usepackage{helvet}  
\usepackage{courier}  
\usepackage[hyphens]{url}  
\usepackage{graphicx} 
\usepackage{natbib}  
\usepackage{caption} 
\DeclareCaptionStyle{ruled}{labelfont=normalfont,labelsep=colon,strut=off} 
\usepackage{algorithm}
\usepackage{algorithmic}

\usepackage{times}
\usepackage{epsfig}
\usepackage{graphicx}
\usepackage{amsmath}
\usepackage{amssymb}
\usepackage{CJKutf8}
\usepackage{adjustbox}
\usepackage{subfigure}
\newcommand{\norm}[1]{\left\lVert#1\right\rVert}
\usepackage{multirow}
\usepackage{hhline}
\usepackage{makecell}
\usepackage{xcolor}
\usepackage{color}

\usepackage{newfloat}
\usepackage{listings}

\floatstyle{ruled}
\newfloat{listing}{tb}{lst}{}
\floatname{listing}{Listing}

\pdfoutput=1

\title{Joint 3D Object Detection and Tracking  Using Spatio-Temporal Representation of Camera Image and LiDAR Point Clouds}

\author{
    Junho Koh\textsuperscript{\rm 1}\equalcontrib,
    Jaekyum Kim\textsuperscript{\rm 1}\equalcontrib,
    Jin Hyeok Yoo\textsuperscript{\rm 1}
    Yecheol Kim\textsuperscript{\rm 1}
    Dongsuk Kum\textsuperscript{\rm 2}
    Jun Won Choi\textsuperscript{\rm 1}
}
\affiliations{
    \textsuperscript{\rm 1}Hanyang University\\
    \textsuperscript{\rm 2}Korea Advanced Institute of Science and Technology (KAIST)\\
    \{jhkoh, jkkim, jhyoo, yckim\}@spa.hanyang.ac.kr, dskum@kaist.ac.kr, junwchoi@hanyang.ac.kr
}

\usepackage{bibentry}

\begin{document}
\pdfoutput=1

\maketitle

\begin{abstract}
In this paper, we propose a new joint object detection and tracking (JoDT) framework for 3D object detection and tracking based on camera and LiDAR sensors.
The proposed method, referred to as 3D DetecTrack, enables the detector and tracker to cooperate to generate a spatio-temporal representation of the camera and LiDAR data, with which 3D object detection and tracking are then performed. 
The detector constructs the spatio-temporal features via the weighted temporal aggregation of the spatial features obtained by the camera and LiDAR fusion.
Then, the detector reconfigures the initial detection results using information from the tracklets maintained up to the previous time step. 
Based on the spatio-temporal features generated by the detector, the tracker associates the detected objects with previously tracked objects using a graph neural network (GNN). We devise a fully-connected GNN facilitated by a combination of rule-based edge pruning and attention-based edge gating, which exploits both spatial and temporal object contexts to improve tracking performance. 
The experiments conducted on both KITTI and nuScenes benchmarks demonstrate that the proposed 3D DetecTrack achieves significant improvements in both detection and tracking performances over baseline methods and achieves state-of-the-art performance among existing methods through collaboration between the detector and tracker.
\end{abstract}

\section{Introduction}
Multiple object tracking (MOT) based on sensor measurements (e.g., camera and LiDAR) is  essential for machine perception tasks in robotics and autonomous driving applications. The traditional approach  is the {\it tracking-by-detection} strategy, which detects objects based on a single snapshot of sensor measurements and temporally links the detection results over multiple snapshots. In this approach, detection and tracking are considered independent tasks and thus have been studied separately by different research communities.

Although the {\it tracking-by-detection} approach has been shown to be effective in numerous studies, the following question arises: {\it if our end goal is to identify moving objects based on the sequence of measurements received from the sensors, would it not be beneficial to jointly design and optimize both detectors and trackers to improve the performance of both?}

Several works relevant to JoDT have been reported in the literature \cite{fairmot, jde, gsdt,  3dt, cdt, kieritz2018joint,  trackrcnn, jrmot}. 
The previous JoDT methods can be categorized into two approaches.
The first approach integrates the re-identification network for object association into the detector and jointly optimizes them end-to-end. 
This approach was mostly developed for 2D MOT, including TrackRCNN \cite{trackrcnn}, JDE \cite{jde}, FairMOT \cite{fairmot}, RetinaTrack \cite{retinatrack}, and GSDT \cite{gsdt}. The second approach exploits the intermediate features extracted by the detector for MOT. The appearance cues and motion context identified by the detector were used to perform tracking. 
The 2D MOT methods in this category include MPNTrack \cite{mpntrack}, RNN tracker \cite{kieritz2018joint}, CDT \cite{cdt}, PredNet \cite{munjal2020joint}, and Chained-Tracker \cite{ctracker}. The JoDT methods for 3D MOT include the mono3DT \cite{3dt} and JRMOT\cite{jrmot}.

In this paper, we propose a new JoDT method, referred to as {\it 3D DetecTrack}, which performs  3D MOT based on the sequence of the camera images in conjunction with LiDAR point clouds. In our framework, the detector and tracker cooperate to utilize spatio-temporal information to perform 3D object detection and tracking.

First, the detector generates spatio-temporal features by combining the spatial features produced by the camera and LiDAR sensor fusion over time. This operation creates features for detecting objects with higher accuracy.  
The detection step in 3D DetecTrack also exploits the {\it tracklet}  maintained by the tracker  to improve detection accuracy. 
The region proposal network (RPN) calibrates an objectness score based on the intersection of union (IoU) between the  anchor box and its nearest tracklet box. Then, refinement network aggregates the instance-level features from adjacent frames and adjusts the classification scores in a manner similar to RPN.
This design is inspired by the intuition that the tracklets provide useful cues, which can increase the confidence score for detecting objects.
 
Next, the tracking step in 3D DetecTrack utilizes the spatio-temporal features generated by the detector in the object association task.  According to the 3D bounding boxes provided by the detector, the tracker pools the object features from the temporally aggregated features and point-encoded features. Then, it associates the detected objects with those in the tracklet through a graph neural network (GNN). We consider a fully-connected GNN model to represent both spatial and temporal relations between the objects. However, this approach causes the connections in the graph to become excessively dense, which makes the convergence of the GNN slower. To address this issue, we employ both rule-based edge pruning and attention-based edge gating.  Rule-based edge pruning removes the edges of a graph based on the distance between the objects comprising a pair. Attention-based edge gating learns to weight the edges of the graph depending on the input. Attending to only the critical edges of the graph enables the GNN to operate faster and more accurate. The detector and tracker are trained in an end-to-end fashion and the loss terms related to both tasks are minimized simultaneously.
The contributions of this paper are summarized as follows

\begin{itemize}
    
    \item We propose a novel JoDT method in which the detector and tracker can collaborate to obtain the spatio-temporal representation of multi-sensor data for 3D MOT.

    \item We design a cooperative JoDT model that reconfigures the outputs of both the RPN and the refinement network based on the information inferred from the tracked objects  and also utilizes the spatio-temporal features formed by the detector for object association.
    
    \item We devise a spatio-temporal gated GNN (SG-GNN), which adopts both rule-based pruning and attention-based edge gating to improve the tracking performance.

    \item We evaluate the performance of the proposed 3D DetecTrack on both the KITTI and nuScenes datasets. Our experiments demonstrate that  the proposed 3D DetecTrack achieves dramatic performance improvements over the baselines and outperforms existing 3D MOT methods on both datasets.

\end{itemize}

\section{Related Work}
\begin{figure*}[t]
	\centering
        \centerline{\includegraphics[width=1.0\textwidth]{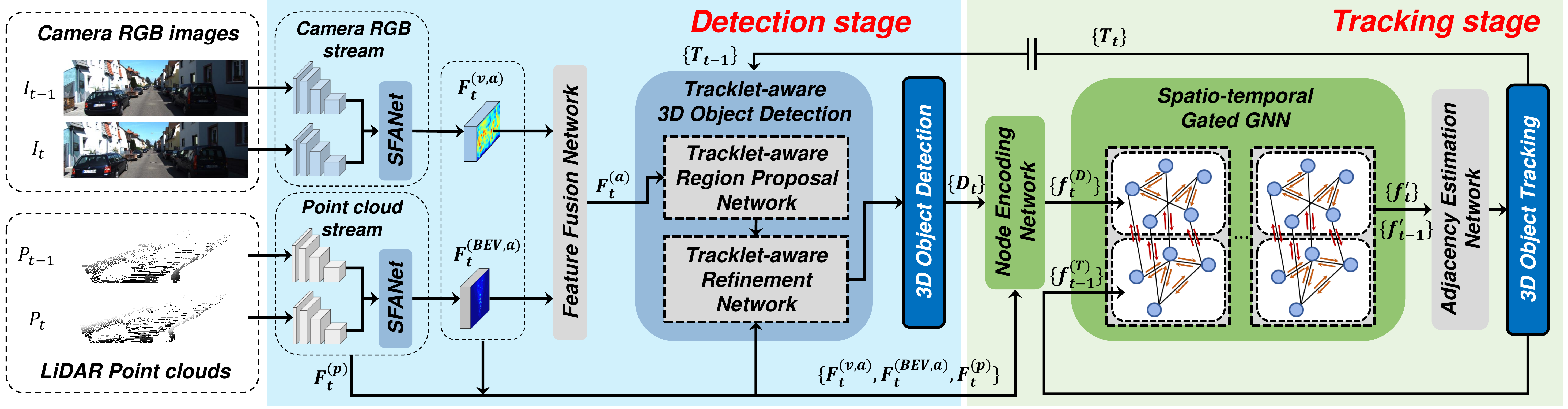}}
    	\caption {\textbf{Overall architecture of the proposed 3D DetecTrack}: 3D DetecTrack effectively utilizes the spatio-temporal information to perform JoDT.  The camera and LiDAR features are temporally aggregated via SFANet. Trk-RPN and Trk-RefNet reconfigure the detection output using the tracklet from the tracker. Using the RoI-aligned features produced by the detector, the objects are associated using SG-GNN. SG-GNN exploits spatio-temporal relations among objects to perform object association. Finally, the affinity matrix is calculated based on the output of SG-GNN. Our proposed network is trained in an end-to-end manner.}
	\label{overall}
\end{figure*} 

\subsection{3D Object Detection and Multi Object Tracking}

3D object detection methods can be divided into two categories: (i) LiDAR-only and (ii) sensor fusion-based methods. The LiDAR-only 3D detectors encode point clouds using the PointNet \cite{pointnet,pointnet++} and detect objects by applying a detection-head network.
These methods include PointRCNN \cite{pointrcnn}, Part A$^2$ \cite{parta2}, STD \cite{std},  3DSSD \cite{3dssd}, VoxelNet \cite{voxelnet}, SECOND \cite{second}, PointPillar \cite{pointpillar}, and CIA-SSD \cite{ciassd}.
To overcome the limitations of the LiDAR-only approach, many sensor fusion-based 3D object detection methods have been developed, which enhance object features by combining intermediate features obtained from the camera image and LiDAR point clouds. 
These methods include ContFuse \cite{contfuse}, MMF \cite{mmf}, CLOCs \cite{clocs}, and 3D-CVF \cite{3dcvf}.
In our method, we adopt a 3D-CVF \cite{3dcvf} as the baseline detector.

Numerous 3D MOT methods follow the {\it tracking-by-detection} paradigm \cite{ab3dmot, maha, gnn3dmot, flowmot, mmmot, fantrack}, in which the objects are first detected in 3D, the tracker then associates the detected objects. Since a simple but effective baseline 3D MOT method, AB3DMOT was proposed in \cite{ab3dmot}. Various 3D MOT methods have been proposed including mmMOT \cite{mmmot}, FANTrack \cite{fantrack}, and GNN3DMOT \cite{gnn3dmot}.
These methods do not consider the co-optimization of detection and tracking, limiting their potential for further performance gains.

\subsection{Joint Object Detection and Tracking}

Various JoDT methods have been proposed in \cite{fairmot, jde, gsdt, ctracker, 3dt, jrmot, cdt, kieritz2018joint, ke2019multi, munjal2020joint}.
In the early phase, the idea of JoDT was adopted for 2D MOT. CDT \cite{cdt} restored undetected objects by examining forward and backward tracing of tracking results. 
JDE \cite{jde} and FairMOT \cite{fairmot} incorporated a re-identification model of the tracker into the detector. 
GSDT \cite{gsdt} performed simultaneous object detection and tracking using GNN. 

Recently, JoDT has been extended to 3D MOT. Mono3DT \cite{3dt} enhanced the performance of monocular camera-based 3D object detection by tracking moving objects across frames via occlusion-aware association and depth-ordering matching.
JRMOT \cite{jrmot} combined re-identification, detection and tracking steps into a joint probabilistic data association framework.
The proposed 3D DetecTrack differs from the aforementioned methods in that the detector and the tracker cooperatively generate strong spatio-temporal features, which are then used to perform 3D object detection and tracking.

\section{Proposed Method}
In this section, we present the details of the proposed  \textit{3D DetecTrack} method.

\subsection{Overview}
The overall architecture of the proposed 3D DetecTrack is depicted in Figure \ref{overall}. The proposed JoDT consists of i) detection stage and ii) tracking stage. 
We denote the tracklet at the $(t-1)$th time step as $T_{t-1}=(T_{t-1,1},...,T_{t-1,N_T})$, where $N_T$ is the number of tracked objects. The objective of JoDT is to produce the detection results $\mathcal{D}_t=(\mathcal{D}_{t,1},...,\mathcal{D}_{t,N_D})$ and the tracklet $T_{t}$ given $T_{t-1}$ and the sensor measurements $Y_t$ acquired at the $t$th time step, where $N_D$ is the number of detected objects. Note that $N_T$ and $N_D$ can vary over time.

\textbf{Detection stage:} We adopt the two-stage detection model, 3D-CVF \cite{3dcvf} as our baseline 3D object detector. We choose the 3D-CVF because it can generate both camera-view 2D features and bird's eye view (BEV) LiDAR features. In our work, we modify the 3D-CVF to produce spatio-temporal features. 
Let ($F_{t-1}^{(v)}, F_{t}^{(v)}$) be the camera-view features extracted from the images ($I_{t-1}, I_{t}$) via the shared CNN backbone network. Similarly, let ($F_{t-1}^{(BEV)}, F_{t}^{(BEV)}$) be the BEV features obtained by the voxelization process followed by the 3D CNN.
The {\it spatio-temporal feature aggregation network (SFANet)} aggregates two concatenated features $(F_{t-1}^{(v)}, F_{t-1}^{(BEV)})$ and $(F_{t}^{(v)}, F_{t}^{(BEV)})$ to produce the spatio-temporal features $(F_{t}^{(v,a)}, F_{t}^{(BEV,a)})$. Because these two features are temporally correlated but not exactly identical, they should be combined in different proportions, depending on their relevance to the end task. 
For this goal, we employ a {\it gated attention mechanism}, which adaptively balances the contributions of $(F_{t-1}^{(v)}, F_{t-1}^{(BEV)})$ and $(F_{t}^{(v)}, F_{t}^{(BEV)})$ by multiplying the learnable weight maps $A_{t-1}$ and $A_{t}$. This operation will be described in detail later.
The detector then fuses the camera features ${F}_t^{(v,a)}$ and LiDAR features ${F}_t^{(BEV,a)}$ into ${F}_t^{(a)}$, following the procedure of 3D-CVF \cite{3dcvf}.

Based on the features ${F}_t^{(a)}$ produced by SFANet, the RPN predicts the object box and objectness scores for each anchor box. Subsequently, the refinement stage refines the box coordinates and computes the classification scores. Our 3D DetecTrack enhances the RPN and refinement stages by utilizing the information obtained from the tracklet $T_{t-1}$.
The Trk-RPN calibrates the objectness score based on the IoU between each anchor and its nearest tracklet in the BEV domain. The objects maintained in the tracklet are supposed to increase the likelihood of objects being detected in the vicinity, which implies that the tracker assists with the detection task. 
The Trk-RefNet aggregates the instance-level features at time $t$ and $t-1$ based on the cosine similarity-based attention and adjusts the classification score based on IoU between the anchor and the nearest tracklet. 

\textbf{Tracking stage:} The tracking stage associates the detection results $\mathcal{D}_t$ with the tracklet $T_{t-1}$ based on the spatio-temporal features obtained in the detection stage. The 3D boxes for the detected objects in $\mathcal{D}_t$ are projected into the BEV domain to produce 2D boxes. Then, 2D RoI pooling is performed for the $i$th object to extract the RoI-aligned features, $f_{t,i}^{(v,a)}$ and $f_{t,i}^{(BEV,a)}$ from $F_{t}^{(v,a)}$ and $F_{t}^{(BEV,a)}$, respectively. In addition, the tracker pools the point-encoded features $f_{t,i}^{(p)}$ via 3D-RoI alignment method proposed in \cite{3dcvf}. These features are concatenated as $f_{t,i}^{(\mathcal{D})}  = (f_{t,i}^{(v,a)},f_{t,i}^{(BEV,a)},f_{t,i}^{(p)})$.  Similarly, a similar feature pooling procedure is applied for the $i$th object in the tracklet $T_{t-1}$. This yields the concatenated features $f_{t-1,i}^{(T)}  = (f_{t-1,i}^{(v,a)},f_{t-1,i}^{(BEV,a)},f_{t-1,i}^{(p)})$. Two feature groups $f_{t,i}^{(\mathcal{D})}$ and  $f_{t-1,i}^{(T)}$ serve for the discriminative features used for object association.

Two features $(f_{t,1}^{(\mathcal{D})},...,f_{t,N_D}^{(\mathcal{D})})$ and $(f_{t-1,1}^{(T)},...,f_{t-1,N_T}^{(T)})$ are fed into the SG-GNN. These features are represented by nodes in the graph. The SG-GNN associates the objects in $\mathcal{D}_t$ with those in $T_{t-1}$ by matching $(f_{t,1}^{(\mathcal{D})},...,f_{t,N_D}^{(\mathcal{D})})$ and $(f_{t-1,1}^{(T)},...,f_{t-1,N_T}^{(T)})$.
We consider a fully connected graph to model the spatial and temporal relations among the objects in $\mathcal{D}_t$ and $T_{t-1}$.  
Because these dense connections can lead to unnecessary feature exchanges between nodes, we devise rule-based pruning and attention-based edge gating to improve SG-GNN. 
Finally, the SG-GNN produces an affinity matrix based on the pairwise association score for all edges connecting the nodes from $\mathcal{D}_t$ and those from $T_{t-1}$. The affinity matrix is then processed by the Hungarian algorithm  \cite{hungarian} to output the final tracklet $T_t$. The entire procedure is repeated until the input sequence is complete.

\begin{figure}[t]
    \centering
    \begin{subfigure}[]{\includegraphics[width=.85\columnwidth]{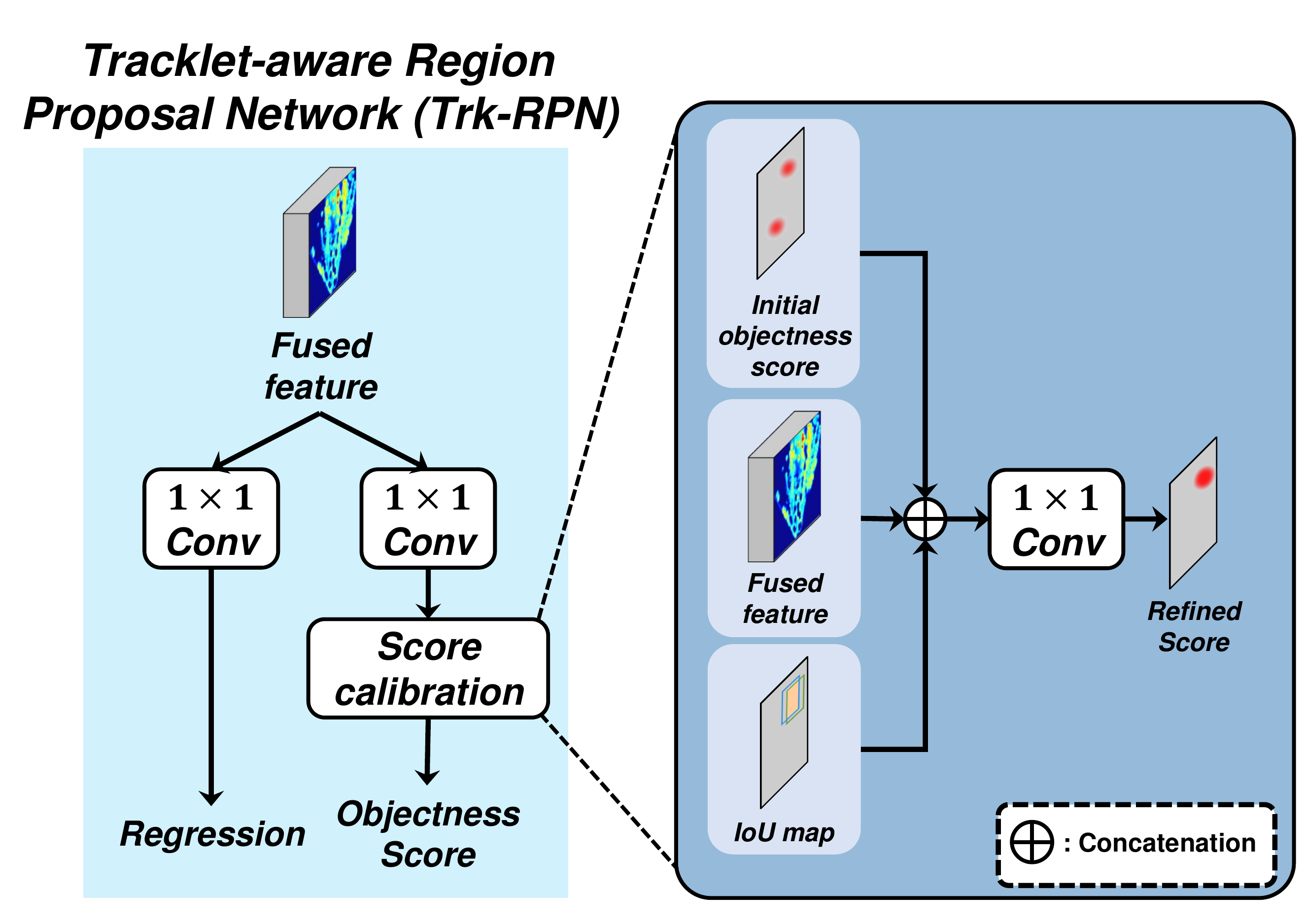}}
    \label{fig:tracklet1}
    \end{subfigure}
    \hspace{0cm}
    \begin{subfigure}[]{\includegraphics[width=.85\columnwidth]{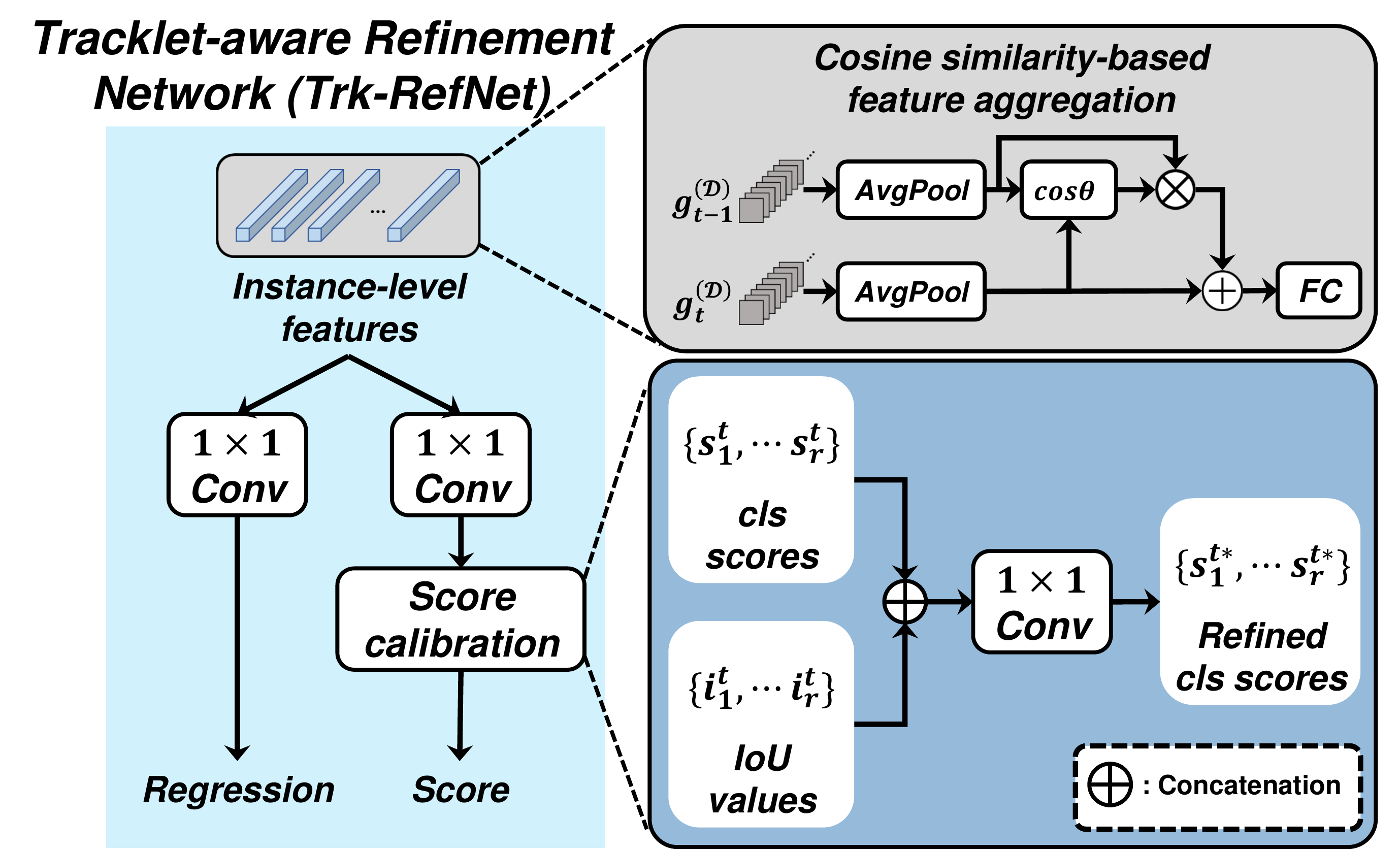}}
    \label{fig:tracklet2}
    \end{subfigure}
    \hspace{0cm}
    \caption {{\bf Structure of (a) Tracklet-aware Region Proposal Network (Trk-RPN) and (b) Tracklet-aware Refinement Network (Trk-RefNet):} 
    Both Trk-RPN and Trk-RefNet utilize the tracklet $T_{t-1}$ to improve the detection performance. 
    }
    \label{fig:tracklet}
\end{figure}

\subsection{Spatio-Temporal Feature Aggregation Network}

The SFANet selectively combines two spatial feature maps obtained in two adjacent time steps. 
 
Suppose that there exist two feature maps ${F}_{t-1}^{(\cdot)}$ and ${F}_{t}^{(\cdot)}$ of the same size $C \times X \times Y$. 
The SFANet applies the weighted aggregation of ${F}_{t-1}^{(\cdot)}$ and ${F}_{t}^{(\cdot)}$ as 
\begin{align}
    {F}_{t}^{(\cdot,a)} = {A}_{t} \otimes {F}_{t}^{(\cdot)}+{A}_{t-1}\otimes {F}_{t-1}^{(\cdot)},
\end{align}
where $\otimes$ denotes pixel-wise multiplication and the attention  maps ${A}_{t}$ and ${A}_{t-1}$ have a size of $1 \times X \times Y $.
The attention maps ${A}_{t}$ and ${A}_{t-1}$ are computed as
\begin{gather}
    {A}_{t} = \sigma(\mathrm{conv_{3\times3}}({F}_{t}^{(\cdot)} \oplus {F}_{t-1}^{(\cdot)}))\\
    {A}_{t-1} = 1-{A}_{t}
\end{gather}
where $\sigma(\cdot)$ is the logistic-sigmoid function, ${\mathrm{ conv_{3\times3}}}$ are the convolutional layers with $3 \times 3$ kernels, and the operation $\oplus$ denotes concatenation.

\subsection{Tracklet-aware 3D Object Detection}
The detailed structures of Trk-RPN and Trk-RefNet are depicted in Figure \ref{fig:tracklet} (a) and (b), respectively.

\textbf{Trk-RPN: } 
Trk-RPN produces $P$ region proposals $\mathcal{R}_1,...,\mathcal{R}_P$ based on the feature maps ${F}_t^{(a)}$ obtained by the SFANet. For each anchor, the initial objectness score and 3D box coordinates are predicted through a 1$\times$1 convolution. The objectness score is calibrated using the IoU between the anchor box and tracklet boxes in $T_{t-1}$ in BEV domain. For each anchor, IoU is obtained as the highest IoU among all tracklet boxes. IoU is set to 0 if the anchor does not overlap with any tracklet box.  The input feature, initial objectness score, and IoU  are concatenated  and passed through the additional feedforward neural network to output the adjusted objectness score.
Because  higher IoU values indicate a higher chance of an anchor being positive, they are expected to increase the objectness score of the anchor.

\textbf{Trk-RefNet: }
Trk-RefNet refines the 3D box prediction and classification score for all region proposals found by Trk-RPN. Based on the $j$th region proposal $\mathcal{R}_j$,  the RoI-aligned features $g_{t,j}^{(v,a)}$ and $g_{t,j}^{(BEV,a)}$ are pooled from $F_{t}^{(v,a)}$ and $F_{t}^{(BEV,a)}$. In addition, 3D-RoI alignment   \cite{3dcvf} is applied to extract the point-encoded features $g_{t,j}^{(p)}$. These features are concatenated as $g_{t,j}^{(\mathcal{D})}  = (g_{t,j}^{(v,a)},g_{t,j}^{(BEV,a)},g_{t,j}^{(p)})$.  Whereas SFANet combines feature maps without any alignment, Trk-RefNet spatially aligns object features before performing feature aggregation. Treating $\mathcal{R}_j$ as an anchor, Trk-RPN predicts the 3D bounding box for the $(t-1)$th time step based on $F_{t-1}^{(a)}$. Then, the concatenated features for the $(t-1)$th time step are obtained as $g_{t-1,j}^{(\mathcal{D})}  = (g_{t-1,j}^{(v,a)},g_{t-1,j}^{(BEV,a)},g_{t-1,j}^{(p)})$. Finally, the instance-level features, $g_{t,j}^{(\mathcal{D})}$ and  $g_{t-1,j}^{(\mathcal{D})}$ are aggregated as
\begin{align}
h_{t,j}   = g_{t,j}^{(\mathcal{D})} + w(g_{t,j}^{(\mathcal{D})},g_{t-1,j}^{(\mathcal{D})}) g_{t-1,j}^{(\mathcal{D})},
\end{align}
$w(g_{t,j}^{(\mathcal{D})},g_{t-1,j}^{(\mathcal{D})})$ denotes the cosine similarity defined as
\begin{align}
    w(g_{t,j}^{(\mathcal{D})},g_{t-1,j}^{(\mathcal{D})})=\frac{ \phi(g_{t,j}^{(\mathcal{D})}) \odot \phi(g_{t-1,j}^{(\mathcal{D})})}{\norm{\phi(g_{t,j}^{(\mathcal{D})})}_2 \norm{\phi(g_{t-1,j}^{(\mathcal{D})})}_2},
\end{align}
where $\phi(\cdot)$ denotes the global average pooling operation, and $\odot$ denotes the inner product operation. The cosine similarity measures the correlation between two detection results and is used to adjust the weight applied to the features at the $(t-1)$th time step.

Trk-RefNet calibrates the classification score by applying the fully-connected layers to the initial classification score concatenated with the IoU between the region proposal and the nearest tracklet box. 

\subsection{Spatio-Temporal Gated Graph Neural Network}
\textbf{Spatio-temporal graph:}  The graph models the relation among the objects in $\mathcal{D}_t$ and $T_{t-1}$. According to the 3D boxes in $\mathcal{D}_t$ and $T_{t-1}$, RoI pooling is performed to obtain the RoI-aligned features $(f_{t,1}^{(\mathcal{D})},...,f_{t,N_D}^{(\mathcal{D})})$ and $(f_{t-1,1}^{(T)},...,f_{t-1,N_T}^{(T)})$. These RoI-aligned features are represented by nodes, and their explicit pairwise relationships are encoded by the edges.  
Previous works \cite{gnn3dmot, gsdt} used a GNN connecting  the nodes at the $t$th time step with those at the $(t-1)$th time step.  In our work, we consider a spatio-temporal graph, which additionally connects the nodes within the nodes at the $t$th time step as well as those within the nodes at the $(t-1)$th time step.
This captures the spatial relation between the objects. Our fully-connected graph is motivated by the idea that understanding the spatial relation between objects will help better associate the objects temporally. However, owing to the dense connections of the graph, a GNN may require unnecessary exchanges of the features, thereby leading to a higher convergence speed.
To address this issue, we apply rule-based edge pruning and attention-based edge gating to SG-GNN, as described below.

\textbf{Rule-based edge pruning:} 
Rule-based edge pruning discards unnecessary edges from a graph based on the spatial distance between objects. Two types of rules are applied. First, the edges connecting the nodes at the $t$th time step  are pruned if  the Euclidean distance between the object centers in the BEV domain is larger than the threshold $L_s$ meter. Similarly, the edges connecting the nodes at the $(t-1)$th time step are pruned according to the same condition. This implies that the SG-GNN only attends to the spatial relation between nearby objects. Second, the edges connecting the nodes between the $t$th time step and the $(t-1)$th time step are pruned if  the Euclidean distance between objects is larger than $L_t$ meter.  This design is justified by the fact that objects that are distant from each other are less likely to be associated.
 These pruning rules can simplify the node connectivity, thereby requiring much fewer iterations.
In our experiments, we set $L_s=15$ meter and $L_t=5$ meter based on an empirical study.

\textbf{Attention-based edge gating:} 
We also apply attention-based edge gating, in which an {\it attention mechanism} is used to attend the model only on the influential edges.
This method adaptively adjusts the weight for the edge based on the similarity of the two features associated with the nodes at both ends. The basic node feature aggregation step follows the rule suggested in \cite{gnn}.  Consider $N$ source nodes $n_{A,1}, ..., n_{A,N}$ and a single target node $n_B$. In the intermediate iteration of GNN, we have the features $f_{A,1},...,f_{A,N}$ and $f_B$ at the nodes $n_{A,1}, ..., n_{A,N}$ and $n_B$, respectively.  The attention weight $a_i$ is applied to the directed edge $E_{n_{A,i} \rightarrow n_B}$ from $n_{A,i}$ to $n_B$ during node feature aggregation. Specifically,  $a_i$ is multiplied by the features $f_{A,i}$ when aggregating the features $f_{A,1},...,f_{A,N}$ at node $n_B$. The attention weight $a_i$ is calculated by 
\begin{align}
s_k &=     \frac{f_{A,k} \circ f_B}{ {\norm{{f}_{A,k}}_2\norm{{f}_{B}}_2} } \\
a_i &= \frac{e^{w \cdot s_i}}{\sum_{k=1}^{N}e^{w \cdot s_k}}
\end{align}
where $\circ$ denotes the dot product, and $w$ is the learnable parameter.
This attention weight reduces the influence of edges in which the two features are not aligned well. This enables the SG-GNN to focus only on the association of  important features.  

\renewcommand{\arraystretch}{0.9}
\begin{table*}[tbh]
\begin{center}
\begin{adjustbox}{width=0.9\textwidth}
\begin{tabular}{c|c|c|c|c|c|c|c}
\Xhline{4\arrayrulewidth}
Method & Input Data & Runtime (ms) & sAMOTA (\%) & AMOTA (\%) & AMOTP (\%) & MOTA (\%) & MOTP (\%) \\\hline
FlowMOT \cite{flowmot} & 3D & - & 90.56 & 43.51 & 76.08 & 85.13 & 79.37 \\
AB3DMOT \cite{ab3dmot} & 3D & 4.7 & 93.28 & 45.43 & 77.41 & 86.24 & 78.43    \\
mmMOT \cite{mmmot} & 2D + 3D & 20 & 70.61 & 33.08 & 72.45 & 74.07 & 78.16   \\
FANTrack \cite{fantrack} & 2D + 3D & 70 & 82.97 & 40.03 & 75.01 & 74.30 & 75.24 \\
GNN3DMOT \cite{gnn3dmot} & 2D + 3D & - & 93.68 & 45.27 & 78.10 & 84.70 & 79.03  \\
Method of \cite{trajectory} & 2D + 3D & - & 92.37 & 44.96 & 76.83 & 84.49 & 78.32     \\
PC-TCNN \cite{pc-tcnn} & 3D & - & 95.44 & 47.64 & - & - & -     \\\hline 
3D-CVF \cite{3dcvf} + mmMOT \cite{mmmot} & 2D + 3D & 20 & 75.22 & 35.50 & 78.06 & 75.62 & 79.73      \\
3D-CVF \cite{3dcvf} + AB3DMOT \cite{ab3dmot} & 2D + 3D & 4.7 & 93.85 & 46.47 & 79.64 & 88.83 & 80.11  \\
Proposed method  & 2D + 3D & 160 (37) & \textbf{96.49} & \textbf{48.87} & \textbf{81.56} & \textbf{91.46} & \textbf{82.24}   \\
\Xhline{4\arrayrulewidth}
\end{tabular}
\end{adjustbox}
\end{center}
\caption{\textbf{3D MOT performance on KITTI tracking \textit {valid} set for \textit{Car} class:}  The number in parentheses  indicates the runtime for the tracking stage only.}
\label{table:3d_mot}
\end{table*}
\renewcommand{\arraystretch}{1}

\renewcommand{\arraystretch}{0.9}
\begin{table*}[hbt!]
\begin{center}
\begin{adjustbox}{width=0.7\textwidth}
\begin{tabular}{c|c|c|c|c|c|c}
\Xhline{3\arrayrulewidth}
Method & mAP & sAMOTA (\%) & AMOTA (\%) & AMOTP (\%) & MOTA (\%) & MOTP (\%) \\\hline
FANTrack \cite{fantrack} & - & 19.64& 2.36& 22.92& 18.60& 39.82\\
mmMOT \cite{mmmot} & -  & 23.93& 2.11& 21.28& 19.82& 40.93\\
GNN3DMOT \cite{gnn3dmot} & - & 29.84& 6.21& 24.02& 23.53& 46.91 \\
AB3DMOT \cite{ab3dmot} & -  & 39.90& 8.94& \textbf{29.67}& 31.40& \textbf{57.54 }\\
Method of \cite{trajectory} & - & 28.96& 11.36& 25.83& 22.81& 41.99\\ \hline 
Baseline & 49.15  & 39.48 & 8.99 & 24.17  & 36.91 & 54.20  \\
Proposed method & \textbf{52.89}  & \textbf{45.60} & \textbf{11.43} & 27.69  & \textbf{43.49} & 55.57  \\
\Xhline{3\arrayrulewidth}
\end{tabular}
\end{adjustbox}
\end{center}
\caption{\textbf{3D MOT performance on nuScenes validation set}}
\label{table:3d_mota_nus}
\end{table*}
\renewcommand{\arraystretch}{1}

\subsection{Object Association} 
After a fixed number of iterations, the SG-GNN ends up with the features $(f'_{t,1}, \cdots, f'_{t,N_D})$ and $(f'_{t-1,1}, \cdots, f'_{t-1,N_T})$ at the nodes.
The affinity matrix $A$ is constructed based on the association score between the objects in $\mathcal{D}_t$ and those in $T_{t-1}$. 
The $(i,j)$th element of $A$ is provided by the association score $A_{ij}$ between $f'_{t-1,i}$ and $f'_{t-1,j}$, which is calculated by
\begin{gather}
    A_{ij} = \sigma(\mathrm{fc}(f'_{t-1,i} \otimes f'_{t,j}))
\end{gather}
where $\mathrm{fc}$ represents fully connected layers with a depth of 3. The association score has a value between 0 and 1  and the affinity matrix $A$ is fed to the Hungarian algorithm \cite{hungarian} to determine the tracklet $T_t$.

\renewcommand{\arraystretch}{0.99}
\begin{table*}[t]
\begin{center}
\begin{adjustbox}{width=0.9\textwidth}
\begin{tabular}{c|c|c|c|c|c|c|c|c|c|c}
\Xhline{4\arrayrulewidth}
 & SG-GNN &  Trk-RPN & Trk-RefNet & SFANet  & AP$_{easy}$ & AP$_{mod.}$ & AP$_{hard}$ & sAMOTA (\%) & AMOTA (\%) & AMOTP (\%) \\\hline
Baseline & & & & & 90.32 & 89.37 & 88.89 & 92.15 & 45.23 & 77.46 \\ \hline
\multirow{3}{*}{Ours} & \checkmark & & & & 90.32 & 89.37 & 88.89 & 94.97 & 47.93 & 80.48 \\
 & \checkmark& \checkmark & & & 97.56 & 89.96 & 89.35 & 95.56 & 48.34 & 80.89\\
 & \checkmark& \checkmark & \checkmark & & 98.41 & 90.32 & 90.08 & 96.02 & 48.56 & 81.16\\
 & \checkmark& \checkmark & \checkmark & \checkmark & \textbf{99.27} & \textbf{91.01} & \textbf{90.83} & \textbf{96.49} & \textbf{48.87} & \textbf{81.56}\\
\Xhline{4\arrayrulewidth}
\end{tabular}
\end{adjustbox}
\end{center}
\caption{\textbf{Ablation study for evaluating SFANet and tracklet-aware 3D object detection:} The ablation study is conducted on KITTI {\it valid} set for {\it Car} class.}
\label{table:ablation_spatio}
\end{table*}
\renewcommand{\arraystretch}{1}

\renewcommand{\arraystretch}{0.99}
\begin{table*}[t]
\begin{center}
\begin{adjustbox}{width=0.9\textwidth}
\begin{tabular}{c|c|c|c|c|c|c|c|c|c}
\Xhline{4\arrayrulewidth}
& SFANet  & Rule-based & Attention-based & \multirow{2}{*}{Runtime (ms)}
& \multirow{2}{*}{sAMOTA (\%)} & \multirow{2}{*}{AMOTA (\%)} & \multirow{2}{*}{AMOTP (\%)} & \multirow{2}{*}{MOTA (\%)} & \multirow{2}{*}{MOTP (\%)}\\

 & + Trk-RPN + Trk-RefNet & edge-pruning& edge gating & & & & & & \\\hline
 
Baseline &  & & & - & 92.15 & 45.23 & 77.46 & 86.72 & 79.52  \\ \hline
\multirow{3}{*}{Ours} & \checkmark & & & 41 & 94.02 & 46.79 & 78.81 & 88.45 & 79.99 \\
 & \checkmark & \checkmark & & \textbf{34} & 95.50 & 48.04 & 81.03 & 90.21 & 81.98\\
 & \checkmark & \checkmark & \checkmark & 37 & \textbf{96.49} & \textbf{48.87} & \textbf{81.56} & \textbf{91.46} & \textbf{82.24} \\
\Xhline{4\arrayrulewidth}
\end{tabular}
\end{adjustbox}
\end{center}
\caption{\textbf{Ablation study for evaluating rule-based edge-pruning and attention-based edge gating:} The ablation study is conducted on KITTI \textit{valid} set for \textit{Car} class. }
\label{table:ablation_gnn}
\end{table*}
\renewcommand{\arraystretch}{1}

\subsection{Loss Function}
We adopt a multi-task loss to train the 3D DetecTrack. The total loss function comprises the detection loss $L_{\rm det}$ and tracking loss $L_{\rm trk}$, i.e.,  $L_{\rm total} = L_{\rm det} + L_{\rm trk}$. 
The detection loss $L_{\rm det}$ is expressed as 
\begin{align}
    L_{\rm det} = L_{\rm rpn} + L_{\rm ref}
\end{align}
where $L_{\rm rpn}$ denotes the RPN loss used to train the network pipeline up to Trk-RPN, and $L_{\rm ref}$ denotes the refinement loss in training Trk-RefNet. 
Following the setup in \cite{3dcvf}, the RPN loss $L_{\rm rpn}$ comprises the focal loss \cite{focal} for the classification task and the Smoothed-L1 loss for the regression task. The refinement loss $L_{\rm ref}$ is defined similarly.

The tracking loss $L_{\rm trk}$ measures the mean squared error (MSE) of the affinity matrix $A$, defined as
\begin{align}
    L_{\rm trk} = \frac{1}{N_D\cdot N_T}\sum_{i=1}^{N_T}\sum_{j=1}^{N_D}(A_{ij}-A^{gt}_{ij})^2.
\end{align}
Note that $A^{gt}$ represents the ground truth (GT) of the affinity matrix $A$. The GT affinity matrix $A^{gt}$ is calculated in two steps. In the first step, the 3D IoU between the output $D_t$ and its GT boxes is calculated and the object ID of the GT box is assigned to the object in $D_t$ if the highest 3D IoU is above 0.5. Comparing the object IDs in $T_{t-1}$ and those assigned to $D_t$, we assign 0 or 1 to the GT affinity matrix $A^{gt}$. 
In the training phase, our 3D DetecTrack model regards the GT for $D_{t-1}$ as the tracklet $T_{t-1}$. 

\section{Experiments}
\subsection{KITTI dataset}
{\bf Dataset and evaluation metrics: }
The KITTI dataset was collected from urban driving scenarios using a single Pointgrey camera and Velodyne HDL-64E LiDAR \cite{kitti}. 
To validate our method,  we split the tracking training dataset evenly into {\it train} set and \textit{valid} set by half, following \cite{gnn3dmot}. Because the KITTI object detection dataset did not contain sequence data, performance was evluated only on the KITTI object tracking dataset.
As a benchmark, we evaluated the 3D MOT performance on the KITTI object tracking {\it valid} dataset.
As 3D MOT performance metrics, we used sAMOTA, AMOTA, and AMOTP metrics \cite{ab3dmot} as well as the standard CLEAR metric \cite{clear}.
We also examine the 2D MOT performance on the KITTI object tracking {\it test} dataset for the {\it Car} category.
For 2D MOT, we used the KITTI evaluation metric \cite{hota} including HOTA, DetA, and AssA. 
Because our method was designed under the JoDT framework, we also evaluated the 3D detection performance. As a performance metric, we used the average precision (AP) with three difficulty levels, i.e., "easy", "moderate", and "hard" as suggested in \cite{kitti}.

{\bf Implementation details: }
Our 3D DetecTrack model was trained in two steps.
In the first step, we trained only the detection stage of the 3D DetecTrack using both the KITTI object detection and tracking datasets. The detection stage requires tracklet $T_{t-1}$ for training. Since the detection dataset does not contain sequence data, the sequence was created by copying the same data.  We trained the detection stage following the setup suggested in \cite{3dcvf}.
In the second step, the detection stage was initialized with the pre-trained model and the entire 3D DetecTrack model was trained end-to-end on the KITTI object tracking dataset. We trained the entire network over $40$ epochs using the ADAM optimizer \cite{adam}. The initial learning rate was set to $10^{-4}$ and decayed by a factor of $0.1$ at the $26$th and $35$th epochs. The weight decay parameter was set to $10^{-4}$, and the mini-batch size was set to $4$. 

\subsection{nuScenes dataset}
{\bf Dataset and evaluation metrics: }
The nuScenes dataset is a large scale autonomous driving dataset which contains more than 1,000 driving scenarios. The dataset was collected using six multi-view cameras, 32-channel LiDAR, and 360-degree object annotations. 
We evaluated 3D MOT performance for 7 categories (bicycle, bus, car, motorcycle, pedestrian, trailer and truck), as a subset of the detection categories in \cite{nuscenes}.
We also evaluated the 3D detection performance for 10 categories, including barrier, construction vehicle, and traffic cone as well as the aforementioned 7 categories. 
As 3D MOT performance metrics, we used the standard CLEAR metric \cite{clear} and the sAMOTA, AMOTA, and AMOTP metrics \cite{ab3dmot}. We used the nuScenes detection score (NDS) \cite{nuscenes} as a 3D detection performance metric.

{\bf Implementation details: }
We trained the 3D DetecTrack using a similar procedure similar to that used for the KITTI benchmark.
We trained the entire model over 20 epochs. The initial learning rate was set to $10^{-4}$ and decayed by a factor of 0.1 at the $13$th and $17$th epochs. The rest of configurations were the same as those for KITTI.

\subsection{Experimental Results}
 
\textbf{Performance on KITTI:}
Table \ref{table:3d_mot} presents the 3D MOT performance and runtime evaluated on the KITTI tracking {\it valid} set. The KITTI benchmark provides a test dataset for 2D MOT, but not for 3D MOT; therefore, we strictly followed the 3D MOT evaluation procedure presented in \cite{ab3dmot}. 
We compared our 3D DetecTrack with several outstanding 3D MOT methods including FlowMOT \cite{flowmot}, AB3DMOT \cite{ab3dmot}, mmMOT \cite{mmmot}, FANTrack \cite{fantrack}, GNN3DMOT \cite{gnn3dmot}, the method of \cite{trajectory}, and PC-TCNN \cite{pc-tcnn}. 
These 3D MOT methods adopt different 3D detectors. Thus, as shown in Table \ref{table:3d_mot}, we also evaluate the performance of  mmMOT \cite{mmmot} and AB3DMOT \cite{ab3dmot} combined with the vanilla 3D-CVF  to compare the ability of the trackers only. 
Table \ref{table:3d_mot} shows that the proposed method outperforms the existing 3D MOT methods by a significant margin for all MOT metrics considered. 
In particular, our 3D DetecTrack performs better than the current state-of-the-art method, PC-TCNN.  
The proposed 3D DetecTrack also achieves better performance than AB3DMOT \cite{ab3dmot} and mmMOT \cite{mmmot} when 3D-CVF is used as a 3D detector. 

We also evaluate the 2D MOT performance on the KITTI object tracking {\it test} set (refer to the official evaluation benchmark on the KITTI leaderboard).
Due to space concerns, we provide the 2D MOT performance in the Appendix.
2D MOT results are obtained by projecting 3D bounding boxes to the camera domain and generating 2D bounding boxes enclosing the projected coordinates.
Although the proposed 3D DetecTrack is not originally designed for 2D MOT task, it achieves comparable performance to that of the current state-of-the-art 2D MOT methods.
Note that the runtime of the tracker in the proposed method (37\textit{ms}) is comparable to that of the existing tracking methods.

\textbf{Performance on nuScenes:}
Table \ref{table:3d_mota_nus} presents the 3D MOT performance on nuScenes validation set. The baseline algorithm uses the original 3D-CVF and the vanilla fully-connected GNN. The proposed approach improves on the baseline  by 3.74\% in mAP and 6.12\%, 2.44\% and 3.52\% in sAMOTA, AMOTA, and AMOTP, respectively. The 3D DetecTrack achieves the best performance among the candidates in the sAMOTA, AMOTA, and MOTA metrics. The per-class performance on both 3D detection and MOT tasks is provided in the Appendix. This shows that the proposed method achieves a particularly remarkable performance gain on bicycle, motorcycle, bus, and truck categories. 

\subsection{Ablation Study}
In this section, we present an ablation study conducted to validate the contributions of the proposed design to our 3D DetecTrack. Experiments were conducted on the KITTI \textit{valid} set. 
Our baseline was chosen as a combination of the original 3D-CVF and the baseline GNN.
Table \ref{table:ablation_spatio} presents how much the components of the 3D DetecTrack, SG-GNN, SFANet, Trk-RPN, and Trk-RefNet improve the performance on both 3D detection and MOT tasks. 
When the SG-GNN is added  to the baseline, both temporal and spatial object context is exploited for object association, improving the sAMOTA performance by 2.82\%. Both Trk-RPN and Trk-RefNet improve both the detection and MOT performance by utilizing the information from the tracklets. Trk-RPN yields a 0.59\% gain in  ${\rm AP}_{mod.}$ detection performance and a 0.59\% gain in sAMOTA MOT performance. Trk-RefNet achieves a 0.36\% gain in ${\rm AP}_{mod.}$ and a 0.46\% gain in sAMOTA. Totally, both  Trk-RPN and Trk-RefNet achieve 0.95\% and 1.05\% gains in ${\rm AP}_{mod.}$ and sAMOTA metrics, respectively.
SFANet also improves the performance by 0.69\% in ${\rm AP}_{mod.}$ and 0.47\%  in sAMOTA by aggregating adjacent feature maps over time. 
Combining all the ideas, the proposed method improves ${\rm AP}_{mod.}$ performance by 1.64\% and sAMOTA performance by 4.34\% over the baseline, which appears to be substantial.

Table \ref{table:ablation_gnn} analyzes the performance gains achieved by the rule-based edge pruning and attention-based edge gating. We evaluate the performance gains achieved by adding each idea to a fully-connected GNN.  
The rule-based edge pruning offers a 1.48\% improvement in sAMOTA by removing unnecessary connections from the fully-connected graph. The attention-based edge gating weights the GNN features according to their importance, which offers additional gain of 0.88\% in sAMOTA. Totally, the combination of the two increases sAMOTA by 2.36\% over the baseline GNN. 
We also analyze the runtime of the SG-GNN. 
While the rule-based pruning method reduces the computation time, the attention-based gating method increases the computation time for better performance. Overall, the SG-GNN reduces the runtime of the baseline GNN by 10\%.

\section{Conclusions}
In this paper, we proposed a novel 3D JoDT method based on camera and LiDAR sensor fusion. In our framework, the detector and tracker work together to jointly optimize detection and object association tasks using spatio-temporal features. 
The detector enhances the object features by applying the weighted temporal feature aggregation to both the camera and LiDAR features. The detector uses the tracklet obtained by the tracker to reconfigure the initial outputs of the detector. The tracker uses the spatio-temporal features delivered by the detector for object association. We also devised the SG-GNN, which efficiently matches the objects on the spatio-temporal graph using a combination of rule-based edge pruning and attention-based edge gating. Our evaluation conducted on the KITTI and nuScenes datasets demonstrated that the 3D DetecTrack achieved a significant performance gain over the baseline and achieved state-of-the-art performance in some MOT evaluation categories.

\newpage
\section{Acknowledgements}
This work was partly supported by the Institute of Information \& Communications Technology Planning \& Evaluation (IITP) grant funded by the Korea government (MSIT) (No. 2020-0-01373, Artificial Intelligence Graduate School Program (Hanyang University)) and the National Research Foundation of Korea (NRF) grant funded by the Korea government (MSIT) (No. 2020R1A2C2012146).

\bibliography{aaai22}

\begin{thebibliography}{43}
\providecommand{\natexlab}[1]{#1}

\bibitem[{Baser et~al.(2019)Baser, Balasubramanian, Bhattacharyya, and
  Czarnecki}]{fantrack}
Baser, E.; Balasubramanian, V.; Bhattacharyya, P.; and Czarnecki, K. 2019.
\newblock Fantrack: {3D} multi-object tracking with feature association
  network.
\newblock In \emph{IEEE Intelligent Vehicles Symposium (IV)}, 1426--1433.

\bibitem[{Bernardin and Stiefelhagen(2008)}]{clear}
Bernardin, K.; and Stiefelhagen, R. 2008.
\newblock Evaluating multiple object tracking performance: the clear mot
  metrics.
\newblock \emph{EURASIP Journal on Image and Video Processing}, 2008: 1--10.

\bibitem[{Bras{\'o} and Leal-Taix{\'e}(2020)}]{mpntrack}
Bras{\'o}, G.; and Leal-Taix{\'e}, L. 2020.
\newblock Learning a neural solver for multiple object tracking.
\newblock In \emph{Proceedings of the IEEE/CVF Conference on Computer Vision
  and Pattern Recognition (CVPR)}, 6247--6257.

\bibitem[{Caesar et~al.(2020)Caesar, Bankiti, Lang, Vora, Liong, Xu, Krishnan,
  Pan, Baldan, and Beijbom}]{nuscenes}
Caesar, H.; Bankiti, V.; Lang, A.~H.; Vora, S.; Liong, V.~E.; Xu, Q.; Krishnan,
  A.; Pan, Y.; Baldan, G.; and Beijbom, O. 2020.
\newblock {Nuscenes}: {A} multimodal dataset for autonomous driving.
\newblock In \emph{Proceedings of the IEEE/CVF conference on Computer Vision
  and Pattern Recognition (CVPR)}, 11621--11631.

\bibitem[{Chiu et~al.(2020)Chiu, Prioletti, Li, and Bohg}]{maha}
Chiu, H.-k.; Prioletti, A.; Li, J.; and Bohg, J. 2020.
\newblock Probabilistic {3D} multi-object tracking for autonomous driving.
\newblock \emph{arXiv preprint arXiv:2001.05673}.

\bibitem[{Geiger, Lenz, and Urtasun(2012)}]{kitti}
Geiger, A.; Lenz, P.; and Urtasun, R. 2012.
\newblock Are we ready for autonomous driving? the {KITTI} vision benchmark
  suite.
\newblock In \emph{Proceedings of the IEEE/CVF conference on Computer Vision
  and Pattern Recognition (CVPR)}, 3354--3361.

\bibitem[{Hu et~al.(2019)Hu, Cai, Wang, Lin, Sun, Krahenbuhl, Darrell, and
  Yu}]{3dt}
Hu, H.-N.; Cai, Q.-Z.; Wang, D.; Lin, J.; Sun, M.; Krahenbuhl, P.; Darrell, T.;
  and Yu, F. 2019.
\newblock Joint monocular {3D} vehicle detection and tracking.
\newblock In \emph{Proceedings of the IEEE/CVF International Conference on
  Computer Vision (ICCV)}, 5390--5399.

\bibitem[{Ke et~al.(2019)Ke, Zheng, Chen, Yan, and Li}]{ke2019multi}
Ke, B.; Zheng, H.; Chen, L.; Yan, Z.; and Li, Y. 2019.
\newblock Multi-object tracking by joint detection and identification learning.
\newblock \emph{Neural Processing Letters}, 50(1): 283--296.

\bibitem[{Kieritz, Hubner, and Arens(2018)}]{kieritz2018joint}
Kieritz, H.; Hubner, W.; and Arens, M. 2018.
\newblock Joint detection and online multi-object tracking.
\newblock In \emph{Proceedings of the IEEE/CVF Conference on Computer Vision
  and Pattern Recognition Workshops (CVPRW)}, 1459--1467.

\bibitem[{Kim and Kim(2016)}]{cdt}
Kim, H.-U.; and Kim, C.-S. 2016.
\newblock {CDT: Cooperative} detection and tracking for tracing multiple
  objects in video sequences.
\newblock In \emph{Proceedings of the European Conference on Computer Vision
  (ECCV)}, 851--867.

\bibitem[{Kingma and Ba(2014)}]{adam}
Kingma, D.~P.; and Ba, J. 2014.
\newblock Adam: A method for stochastic optimization.
\newblock \emph{arXiv preprint arXiv:1412.6980}.

\bibitem[{Kuhn(1955)}]{hungarian}
Kuhn, H.~W. 1955.
\newblock The Hungarian method for the assignment problem.
\newblock \emph{Naval research logistics quarterly}, 2(1-2): 83--97.

\bibitem[{Lang et~al.(2019)Lang, Vora, Caesar, Zhou, Yang, and
  Beijbom}]{pointpillar}
Lang, A.~H.; Vora, S.; Caesar, H.; Zhou, L.; Yang, J.; and Beijbom, O. 2019.
\newblock {PointPillars}: {Fast} encoders for object detection from point
  clouds.
\newblock In \emph{Proceedings of the IEEE/CVF conference on Computer Vision
  and Pattern Recognition (CVPR)}, 12697--12705.

\bibitem[{Liang et~al.(2019)Liang, Yang, Chen, Hu, and Urtasun}]{mmf}
Liang, M.; Yang, B.; Chen, Y.; Hu, R.; and Urtasun, R. 2019.
\newblock Multi-task multi-sensor fusion for {3D} object detection.
\newblock In \emph{Proceedings of the IEEE/CVF conference on Computer Vision
  and Pattern Recognition (CVPR)}, 7345--7353.

\bibitem[{Liang et~al.(2018)Liang, Yang, Wang, and Urtasun}]{contfuse}
Liang, M.; Yang, B.; Wang, S.; and Urtasun, R. 2018.
\newblock Deep continuous fusion for multi-sensor {3D} object detection.
\newblock In \emph{Proceedings of the European Conference on Computer Vision
  (ECCV)}, 641--656.

\bibitem[{Lin et~al.(2017)Lin, Goyal, Girshick, He, and Doll{\'a}r}]{focal}
Lin, T.-Y.; Goyal, P.; Girshick, R.; He, K.; and Doll{\'a}r, P. 2017.
\newblock Focal loss for dense object detection.
\newblock In \emph{Proceedings of the IEEE/CVF International Conference on
  Computer Vision (ICCV)}, 2980--2988.

\bibitem[{Lu et~al.(2020)Lu, Rathod, Votel, and Huang}]{retinatrack}
Lu, Z.; Rathod, V.; Votel, R.; and Huang, J. 2020.
\newblock Retinatrack: Online single stage joint detection and tracking.
\newblock In \emph{Proceedings of the IEEE/CVF conference on Computer Vision
  and Pattern Recognition (CVPR)}, 14668--14678.

\bibitem[{Luiten et~al.(2020)Luiten, Osep, Dendorfer, Torr, Geiger,
  Leal-Taix{\'e}, and Leibe}]{hota}
Luiten, J.; Osep, A.; Dendorfer, P.; Torr, P.; Geiger, A.; Leal-Taix{\'e}, L.;
  and Leibe, B. 2020.
\newblock {HOTA: A} higher order metric for evaluating multi-object tracking.
\newblock \emph{International Journal of Computer Vision (IJCV)}, 1--31.

\bibitem[{Morris et~al.(2019)Morris, Ritzert, Fey, Hamilton, Lenssen, Rattan,
  and Grohe}]{gnn}
Morris, C.; Ritzert, M.; Fey, M.; Hamilton, W.~L.; Lenssen, J.~E.; Rattan, G.;
  and Grohe, M. 2019.
\newblock {Weisfeiler} and leman go neural: {Higher-order} graph neural
  networks.
\newblock In \emph{Proceedings of the AAAI Conference on Artificial
  Intelligence}, 4602--4609.

\bibitem[{Munjal et~al.(2020)Munjal, Aftab, Amin, Brandlmaier, Tombari, and
  Galasso}]{munjal2020joint}
Munjal, B.; Aftab, A.~R.; Amin, S.; Brandlmaier, M.~D.; Tombari, F.; and
  Galasso, F. 2020.
\newblock Joint detection and tracking in videos with identification features.
\newblock \emph{Image and Vision Computing}, 100: 103932.

\bibitem[{Pang, Morris, and Radha(2020)}]{clocs}
Pang, S.; Morris, D.; and Radha, H. 2020.
\newblock CLOCs: Camera-LiDAR object candidates fusion for 3D object detection.
\newblock In \emph{Proceedings of the IEEE/RSJ International Conference on
  Intelligent Robots and Systems (IROS)}, 10386--10393.

\bibitem[{Peng et~al.(2020)Peng, Wang, Wan, Wu, Wang, Tai, Wang, Li, Huang, and
  Fu}]{ctracker}
Peng, J.; Wang, C.; Wan, F.; Wu, Y.; Wang, Y.; Tai, Y.; Wang, C.; Li, J.;
  Huang, F.; and Fu, Y. 2020.
\newblock Chained-tracker: {Chaining} paired attentive regression results for
  end-to-end joint multiple-object detection and tracking.
\newblock In \emph{Proceedings of the European Conference on Computer Vision
  (ECCV)}, 145--161.

\bibitem[{Qi et~al.(2017{\natexlab{a}})Qi, Su, Mo, and Guibas}]{pointnet}
Qi, C.~R.; Su, H.; Mo, K.; and Guibas, L.~J. 2017{\natexlab{a}}.
\newblock Pointnet: {Deep} learning on point sets for {3D} classification and
  segmentation.
\newblock In \emph{Proceedings of the IEEE/CVF conference on Computer Vision
  and Pattern Recognition (CVPR)}, 652--660.

\bibitem[{Qi et~al.(2017{\natexlab{b}})Qi, Yi, Su, and Guibas}]{pointnet++}
Qi, C.~R.; Yi, L.; Su, H.; and Guibas, L.~J. 2017{\natexlab{b}}.
\newblock Pointnet++: {Deep} hierarchical feature learning on point sets in a
  metric space.
\newblock In \emph{Advances in Neural Information Processing Systems
  (NeurIPS)}, 5099--5108.

\bibitem[{Shenoi et~al.(2020)Shenoi, Patel, Gwak, Goebel, Sadeghian,
  Rezatofighi, Mart{\'\i}n-Mart{\'\i}n, and Savarese}]{jrmot}
Shenoi, A.; Patel, M.; Gwak, J.; Goebel, P.; Sadeghian, A.; Rezatofighi, H.;
  Mart{\'\i}n-Mart{\'\i}n, R.; and Savarese, S. 2020.
\newblock Jrmot: A real-time {3D} multi-object tracker and a new large-scale
  dataset.
\newblock In \emph{Proceedings of the IEEE/RSJ International Conference on
  Intelligent Robots and Systems (IROS)}, 10335--10342.

\bibitem[{Shi, Wang, and Li(2019)}]{pointrcnn}
Shi, S.; Wang, X.; and Li, H. 2019.
\newblock {PointRCNN}: {3D} object proposal generation and detection from point
  cloud.
\newblock In \emph{Proceedings of the IEEE/CVF conference on Computer Vision
  and Pattern Recognition (CVPR)}, 770--779.

\bibitem[{Shi et~al.(2019)Shi, Wang, Wang, and Li}]{parta2}
Shi, S.; Wang, Z.; Wang, X.; and Li, H. 2019.
\newblock Part-A2 Net: {3D} part-aware and aggregation neural network for
  object detection from point cloud.
\newblock \emph{arXiv preprint arXiv:1907.03670}.

\bibitem[{Voigtlaender et~al.(2019)Voigtlaender, Krause, Osep, Luiten, Sekar,
  Geiger, and Leibe}]{trackrcnn}
Voigtlaender, P.; Krause, M.; Osep, A.; Luiten, J.; Sekar, B. B.~G.; Geiger,
  A.; and Leibe, B. 2019.
\newblock {MOTS: Multi-object} tracking and segmentation.
\newblock In \emph{Proceedings of the IEEE/CVF Conference on Computer Vision
  and Pattern Recognition (CVPR)}, 7942--7951.

\bibitem[{Wang, Kitani, and Weng(2020)}]{gsdt}
Wang, Y.; Kitani, K.; and Weng, X. 2020.
\newblock Joint object detection and multi-object tracking with graph neural
  networks.
\newblock \emph{arXiv preprint arXiv:2006.13164}, 5.

\bibitem[{Wang et~al.(2020)Wang, Zheng, Liu, Li, and Wang}]{jde}
Wang, Z.; Zheng, L.; Liu, Y.; Li, Y.; and Wang, S. 2020.
\newblock Towards real-time multi-object tracking.
\newblock In \emph{Proceedings of the European Conference on Computer Vision
  (ECCV)}, 107--122.

\bibitem[{Weng et~al.(2020{\natexlab{a}})Weng, Wang, Held, and
  Kitani}]{ab3dmot}
Weng, X.; Wang, J.; Held, D.; and Kitani, K. 2020{\natexlab{a}}.
\newblock {3D} multi-object tracking: A baseline and new evaluation metrics.
\newblock In \emph{Proceedings of the IEEE/RSJ International Conference on
  Intelligent Robots and Systems (IROS)}, 10359--10366.

\bibitem[{Weng et~al.(2020{\natexlab{b}})Weng, Wang, Man, and
  Kitani}]{gnn3dmot}
Weng, X.; Wang, Y.; Man, Y.; and Kitani, K.~M. 2020{\natexlab{b}}.
\newblock {GNN3MOT: Graph} neural network for 3d multi-object tracking with
  {2D-3D} multi-feature learning.
\newblock In \emph{Proceedings of the IEEE/CVF Conference on Computer Vision
  and Pattern Recognition (CVPR)}, 6499--6508.

\bibitem[{Weng, Yuan, and Kitani(2020)}]{trajectory}
Weng, X.; Yuan, Y.; and Kitani, K. 2020.
\newblock Joint {3D} tracking and forecasting with graph neural network and
  diversity sampling.
\newblock \emph{arXiv preprint arXiv:2003.07847}.

\bibitem[{Wu et~al.(2021)Wu, Li, Wen, Li, Fan, and Wang}]{pc-tcnn}
Wu, H.; Li, Q.; Wen, C.; Li, X.; Fan, X.; and Wang, C. 2021.
\newblock Tracklet proposal network for multi-object tracking on point clouds.
\newblock In \emph{Proceedings of the International Joint Conference on
  Artificial Intelligence (IJCAI)}, 1165--1171.

\bibitem[{Yan, Mao, and Li(2018)}]{second}
Yan, Y.; Mao, Y.; and Li, B. 2018.
\newblock Second: {Sparsely} embedded convolutional detection.
\newblock \emph{Sensors}, 18(10): 3337.

\bibitem[{Yang et~al.(2020)Yang, Sun, Liu, and Jia}]{3dssd}
Yang, Z.; Sun, Y.; Liu, S.; and Jia, J. 2020.
\newblock {3DSSD}: Point-based {3D} single stage object detector.
\newblock In \emph{Proceedings of the IEEE/CVF conference on Computer Vision
  and Pattern Recognition (CVPR)}, 11040--11048.

\bibitem[{Yang et~al.(2019)Yang, Sun, Liu, Shen, and Jia}]{std}
Yang, Z.; Sun, Y.; Liu, S.; Shen, X.; and Jia, J. 2019.
\newblock {STD: Sparse-to-dense} {3D} object detector for point cloud.
\newblock In \emph{Proceedings of the IEEE/CVF International Conference on
  Computer Vision (ICCV)}, 1951--1960.

\bibitem[{Yoo et~al.(2020)Yoo, Kim, Kim, and Choi}]{3dcvf}
Yoo, J.~H.; Kim, Y.; Kim, J.; and Choi, J.~W. 2020.
\newblock 3D-CVF: Generating joint camera and lidar features using cross-view
  spatial feature fusion for 3d object detection.
\newblock In \emph{Proceedings of the European Conference on Computer Vision
  (ECCV)}, 720--736.

\bibitem[{Zhai et~al.(2020)Zhai, Kong, Cui, Liu, and Yang}]{flowmot}
Zhai, G.; Kong, X.; Cui, J.; Liu, Y.; and Yang, Z. 2020.
\newblock {FlowMOT}: {3D} multi-object tracking by scene flow association.
\newblock \emph{arXiv preprint arXiv:2012.07541}.

\bibitem[{Zhang et~al.(2019)Zhang, Zhou, Sun, Wang, Shi, and Loy}]{mmmot}
Zhang, W.; Zhou, H.; Sun, S.; Wang, Z.; Shi, J.; and Loy, C.~C. 2019.
\newblock Robust multi-modality multi-object tracking.
\newblock In \emph{Proceedings of the IEEE/CVF International Conference on
  Computer Vision (ICCV)}, 2365--2374.

\bibitem[{Zhang et~al.(2021)Zhang, Wang, Wang, Zeng, and Liu}]{fairmot}
Zhang, Y.; Wang, C.; Wang, X.; Zeng, W.; and Liu, W. 2021.
\newblock Fairmot: On the fairness of detection and re-identification in
  multiple object tracking.
\newblock \emph{International Journal of Computer Vision (IJCV)}, 1--19.

\bibitem[{Zheng et~al.(2021)Zheng, Tang, Chen, Jiang, and Fu}]{ciassd}
Zheng, W.; Tang, W.; Chen, S.; Jiang, L.; and Fu, C.-W. 2021.
\newblock {CIA-SSD:} {Confident} IoU-aware single-stage object detector from
  point cloud.
\newblock In \emph{Proceedings of the AAAI Conference on Artificial
  Intelligence}, 3555--3562.

\bibitem[{Zhou and Tuzel(2018)}]{voxelnet}
Zhou, Y.; and Tuzel, O. 2018.
\newblock Voxelnet: {End-to-end} learning for point cloud based {3D} object
  detection.
\newblock In \emph{Proceedings of the IEEE/CVF conference on Computer Vision
  and Pattern Recognition (CVPR)}, 4490--4499.

\end{thebibliography}

\end{document}